\documentclass{article}

\usepackage{graphicx}
\usepackage{algorithmic}
\usepackage{algorithm}

\begin{document}

\title{Centric Selection: a Way to Tune\\ the Exploration/Exploitation 
Trade-off }

%\numberofauthors{4}
\author{
David Simoncini, S\'ebastien Verel\\
Philippe Collard, Manuel Clergue\\
}

\date{ }

\maketitle
\begin{abstract}

In this paper, we study the exploration / exploitation trade-off 
in cellular genetic algorithms. We define a new selection
scheme, the centric selection, which is tunable and allows controlling
 the selective pressure with a single parameter.
The equilibrium model is used to study the influence of the centric 
selection on the selective pressure and a new model
which takes into account problem dependent statistics  and selective pressure
 in order to deal with the exploration / exploitation
trade-off is proposed: the punctuated equilibria model. Performances on the
 quadratic assignment problem and NK-Landscapes put in evidence an optimal 
exploration / exploitation trade-off on both of the classes of problems. 
The punctuated equilibria model is used to explain these results.

\end{abstract}

\section{Introduction}

The exploration/exploitation trade-off is an important 
issue in evolutionary computation. By tuning the selective pressure on the population, one 
can find an optimal (or near-optimal) tradeoff between exploitation and exploration. 
In cellular Evolutionary Algorithms (cEAs), the population is embedded on a bidimensional toroidal grid and each solution interacts 
with its neighbors thanks to a certain neighborhood. The convergence rate of the algorithm is 
then dependent of the shape and size of the grid and of the neighborhood. 
The smallest symetric neighborhood that can be defined is the well-known Von Neumann neighborhood of radius $1$. It guarantees a slow isotropic diffusion of genetic information 
through the grid. But when solving complex multimodal problems, it is necessary
 to slow 
down even more the propagation speed of the best solution because the algorithm still often 
converges over a local optimum.
  
Our goal in this paper is to establish a relation between the selective pressure on the population and the effects of recombination and mutation operators, in order to 
find an optimal exploration/exploitation trade-off. To do so, we propose a 
new selection scheme able to control the selective pressure and a 
theoretical model which takes into account the effects of stochastic variations
on an optimization problem. In section 2 we define a selection scheme able to tune the selective pressure and present the algorithm used in the experiments. 
In section 3, we analyze the selective pressure with respect to the selection 
method and present a new model which takes into account the stochastic 
variations. In section 4 we present performances on Quadratic assignment problem instances and on NK-Landscapes and we explain the results with the model proposed.

\subsection{Cellular Evolutionary Algorithms}

\begin{figure}[ht!]
\begin{center}
\rotatebox{270}{\includegraphics[width=3.5cm,height=3.5cm]{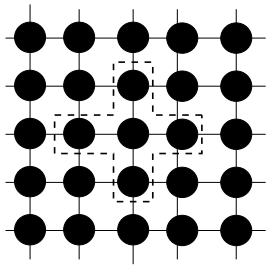}} 
\end{center}
\caption{Representation of a cEA and Von Neumann neighborhood in dashed line.}
\label{cea}
\end{figure}

A cellular Evolutionary Algorithm (cEA) \cite{Whitley93} is an EA in which the population is embedded on a bidimensionnal toroidal grid (see figure \ref{cea}). 
Each cell of the grid contains a solution. Embedding the solutions on a grid 
allows defining a neighborhood between the cells. The most commonly used 
one in cEAs is the Von Neumann neighborhood (shown on figure \ref{cea}). 
At each generation, every cell on the grid is updated by selecting parents in its neighborhood and applying stochastic 
operators such as crossover and mutation. Several strategies exist, synchronous and asynchronous, to update the cells. The small overlapped neighborhoods guarantee the diffusion of solutions through the grid \cite{SpiessensM91}.    
 Such algorithms are especially well suited for complex problems \cite{JongS95}
and are of advantage when dealing with dynamic problems \cite{Tomassini05}.

\subsection{Selective pressure}

One of the main properties that differs between EAs and cEAs is the rate of convergence (propagation speed of the best solution) : It is exponential 
for EAs and quadratic for cEAs. Therefore, the selective pressure on the population is weaker for a cEA than for an EA. Controlling
 the selective pressure is critical since it can avoid premature convergence of the algorithm when solving complex multimodal 
problems. Several parameters related to the selective pressure can prevent the algorithm from getting stuck in a local optimum. 
The topology of the grid, the local neighborhood, the properties of the selection operator are such parameters. By correctly 
tuning these parameters for a given problem, one can find a good exploration/exploitation tradeoff and minimize the risks of premature convergence. Sarma \textit{et al.} established a link between the radius of the neighborhood and the radius of the 
grid : changing this ratio directly affects the selective pressure on the population \cite{SarmaJ96}. Alba \textit{et al.} analyzed 
performances of cEAs with a fixed size neighborhood and different grid shapes. They arrived to the conclusion that 
thin grids are well-suited for complex multi modal problems and large grids are well-suited for simple problems. The main 
explanation is that thinner grids give lower selective pressure \cite{AlbaT00}.
 Takeover times and growth curves analysis are useful to measure the selective pressure on a population, but it is not sufficient to decide of 
a trade-off between exploration and exploitation: it is necessary to include effects of the stochastic variations due to the operators in the analysis. 
 Janson \textit{et al.} proposed 
a hierarchical cEA which allows achieving different levels of exploration /exploitation tradeoff in distinct zones of 
the population simultaneously \cite{Janson06}.

A standard technique to study the induced selective pressure without introducing the
perturbing effect of variation operators
is to let selection be the only active operator, and then monitor the number of best
 solution copies in the population \cite{GoldbergD91}.
The takeover time is the time needed for one single best solution 
to colonize the population with selection as the only active operator. Let $\lambda$ be the size of the population, $t$ the number 
of generations and $N(t)$ the number of best solution copies at generation $t$. 
 The population is initialized with one solution of good fitness and $\lambda - 1$ solutions of null fitness. 
Since no other evolution mechanism but selection takes place, the good fitness solution spreads over the grid. 
The takeover time is then the smaller time $t$ such that $N(t)=\lambda$. Analysing the growth of $N(t)$ as a function of $t$ 
also gives an indication on the selective pressure. It shows the convergence rate of the algorithm when selection is 
the only active operator. When the slope of the growth curve of $N(t)$ is low, the convergence rate is low and 
the takeover time is high. On the other hand, a high slope of the growth curve leads to a short takeover time. 
In the first case, the selective pressure on the population is weaker than in the second case.

\subsection{Growth curves and takeover time models}

Characterizing the growth curves and the takeover time is an important issue in the study of the 
selective pressure \cite{GoldbergD91}. Many models have been proposed to define the behaviour of structured
 population evolutionary algorithms. Sarma and De Jong proposed a logistic model in which the coefficient of the growth 
curve of the best solution is shown to be an inverse exponential of the ratio between radii of the neighborhood 
and the underlying grid \cite{SarmaJ96}. This conclusion was guided by an empirical analysis of the effects 
on the convergence rate and the takeover time of several neighborhood sizes and shapes.
Sprave proposed a hypergraph based model of population structures and a method for the estimation of growth curves and 
takeover times based on the probabilistic diameter of the population \cite{Sprave99}.
Gorges-Schleuter proposed a study about takeover time and growth curves for cellular evolution strategies. She 
obtained a linear model for ring populations and a quadratic model for a torus population structure \cite{Gorges99}.
Several authors wrote about theoretical or empirical models of growth curves and takeover time. Giacobini \textit{et. al.} proposed 
a model for cellular evolutionary algorithms with asynchronous update policies \cite{Giacobini03}. He summarized his results 
and proposed models for synchronous updates \cite{Giacobini05b} that will be evoked later in this paper. 
Alba proposed a model for distributed evolutionary algorithms consisting in the sum of logistic definition of the 
component takeover regimes \cite{Alba04b}. In his paper, he made an interesting review of existing models and 
compared two of them (the logistic model and the hypergraph model) with his newly proposed one. For a detailed 
state of the art of cEAs, see \cite{alba08}.

\section{Centric selection}

In this section we present a new selection scheme for cEAs that allows tuning accurately the selective pressure.

\begin{algorithm}
\begin{algorithmic}
\STATE \textbf{CentricSelection}{index: int, $\beta$: double}
\STATE neighbors $\longleftarrow$ GetNeighborhood(index)
\STATE $candidate_1$ $\longleftarrow$ Select(neighbors, $\beta$)
\STATE $candidate_2$ $\longleftarrow$ Select(neighbors, $\beta$)
\STATE Best($candidate_1$, $candidate_2$)
\end{algorithmic}
\caption{Centric Selection algorithm}
\label{centricAlgo}
\end{algorithm}

The centric selection (CS) idea is to change the probability of selecting the center 
cell of the neighborhood. This scheme allows slowing down the convergence speed while keeping an isotropic diffusion of good solutions through the grid. 
The CS is a determinist tournament selection. But unlike the standard 
deterministic tournament, cells in the neighborhood may have different 
probabilities of being selected for the competition. The anisotropic selection 
\cite{Gecco06} is another selection scheme which modifies the probability 
of selection a cell for a deterministic tournament. With the anisotropic 
selection, the diffusion of solutions is not isotropic, so we propose the CS 
 which is easier to study.  
 We have $p_c = \beta$ the probability 
of selecting the center cell and $p_n = p_s = p_e = p_w = \frac{1}{4} (1-\beta)$ the probability of selecting either north, south, east or west cell. 
When $\beta = \frac{1}{5}$, all cells have the same probability of being selected for the competition: this particular case of CS is the standard binary
 tournament selection. When $\beta=1$, only the center cell can be selected 
for the tournament: in this particular case where the same solution is selected 
two times, the crossover operator is not applied in the cEA. Only mutations 
are applied to the solution, and with an elitist replacement strategy, the 
algorithm behaves as the parallelisation of as many hill climbers as there 
are solutions in the population. 
The CS is described in algorithm \ref{centricAlgo}. The candidates compete in a deterministic tournament returning 
the best one. For each cell on the grid, two parents are selected per 
generation, as we can see in the algorithm \ref{cEAlgo}. Stochastic 
variations operators 
are applied to the parents, generating two children. The replacement 
strategy is elitist: the best child replaces the current solution on the 
grid if it has a better fitness. The use of a temporary grid is necessary 
for a synchronous update of the cells.

\begin{algorithm}
\begin{algorithmic}
\STATE \textbf{cEA}{population: vector, $\beta$: double}

\STATE tempGrid: vector
\WHILE{continue()}
\FOR{$i = 1$ to GridSize}
\STATE $parent_1$ $\longleftarrow$ CentricSelection(i, $\beta$)
\STATE $parent_2$ $\longleftarrow$ CentricSelection(i, $\beta$)

\STATE ($child_1$, $child_2$) $\longleftarrow$ Crossover($parent_1$,$parent_2$);

\STATE Mutate($child_1$)
\STATE Mutate($child_2$)

\STATE tempGrid[i] $\longleftarrow$ Best($population[i]$, $child_1$, $child_2$)
\ENDFOR
\STATE Replace(population, tempGrid)
\ENDWHILE
\end{algorithmic}
\caption{Description of our cEA}
\label{cEAlgo}
\end{algorithm}

\section{Modeling cEAs}

In this section, we present two models of the search dynamic in cEAs. 
In the first one, the Equilibrium Model (EM) we consider that the optimal solution has been found and observe how it colonizes the grid. This model is classical in the studies 
on the selective pressure and one the exploration/exploitation trade-off. 
The informations given by this model are takeover times and best 
solution growth curves. As the stochastic variations operators are not
 taken into
account, the same dynamic occurs in experimental runs when the recombination 
 and mutation operators are ineffective: when the system has reached an equilibrium.
  
 In the second one, we consider that a better solution 
can be found with a certain probability and observe the frequency of 
apparition of this new solution with respect to our algorithm's parameters.
It is a model of the transition between two periods of fitness 
stability. 
We call this new model the Punctuated Equilibria Model (PEM). 

\subsection{Equilibrium model}

In order to measure the selective pressure induced by the CS, we observe what happens when no more solution improvement is possible. 
In this case, crossover and mutation are no longer useful and the evolution process has reached an equilibrium. Hence, we observe the time needed for a single best solution 
to conquer the whole grid, and look at the growth curve obtained and the takeover time. 

\begin{figure}[ht!]
\begin{center}
\rotatebox{270}{\includegraphics[width=6cm,height=6cm]{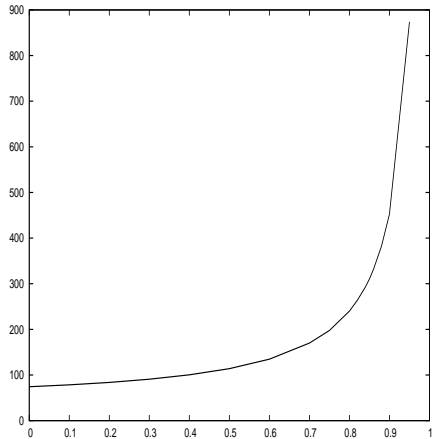}} 
\end{center}
\caption{Average takeover time as a function of $\beta$.}
\label{fig-tak-fuzzy}
\end{figure}

\begin{figure}[ht!]
\begin{center}
\begin{tabular}{c}
\rotatebox{270}{\includegraphics[width=6cm,height=6cm]{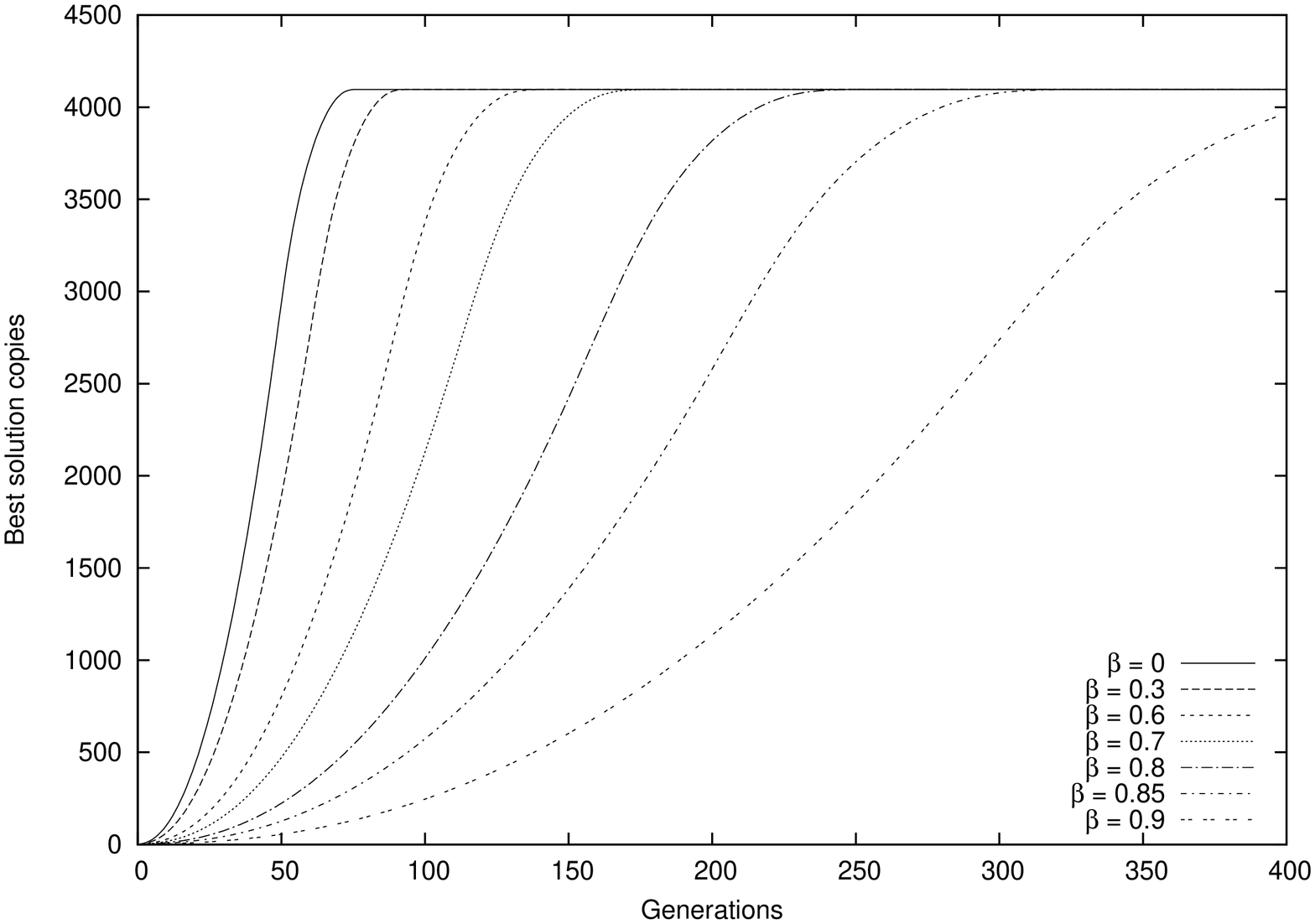}} \\
\rotatebox{270}{\includegraphics[width=6cm,height=6cm]{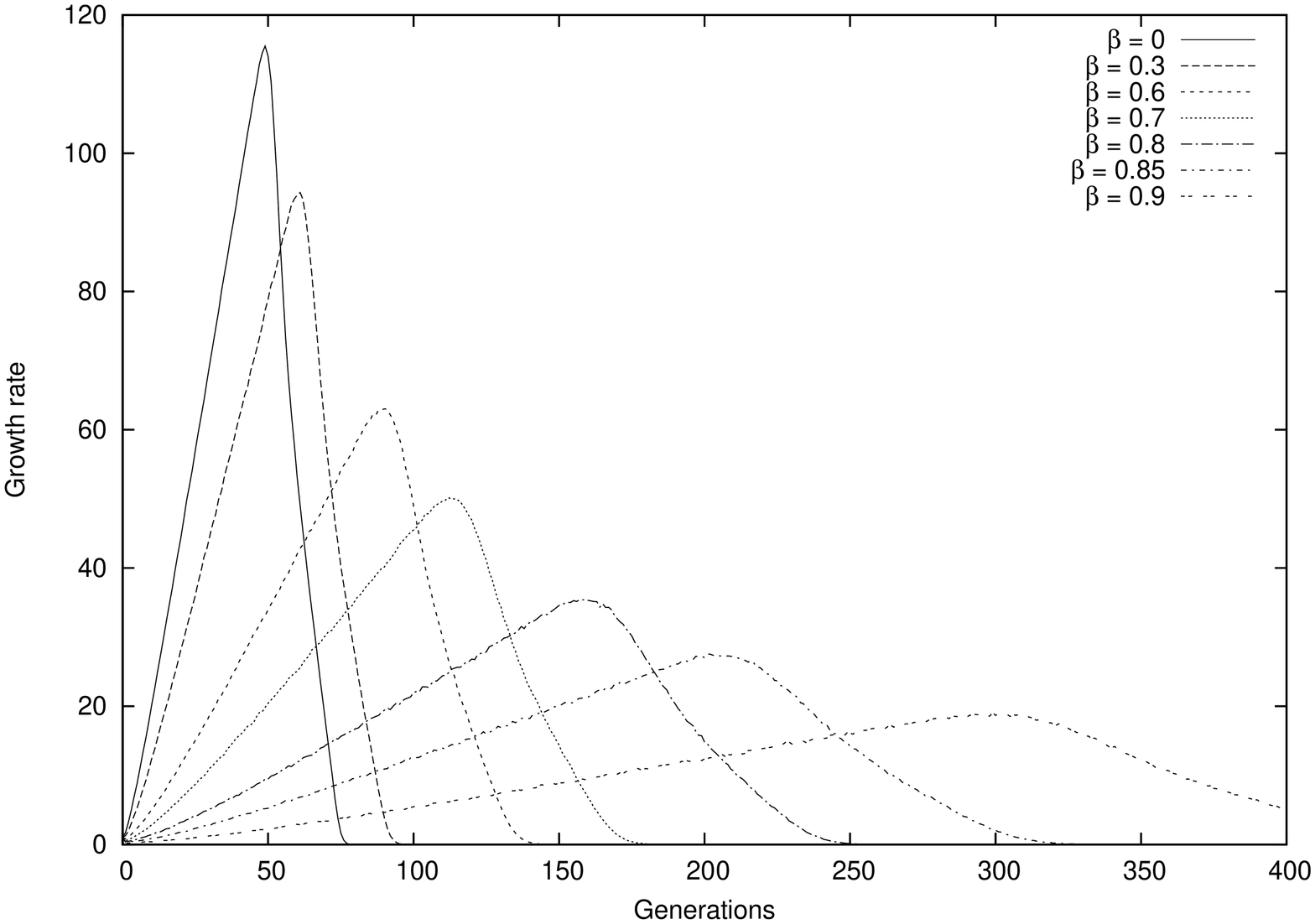}} \\
\end{tabular}
\end{center}
\caption{Growth curves of $N(t)$ (top) and its growth rate (bottom) for different values of $\beta$.}
\label{growthPC}
\end{figure}

We measure the effects of CS on selective pressure by observing these growth curves and takeover times on a square grid of side $64$. 
Figure \ref{fig-tak-fuzzy} shows the takeover time as a function of $\beta$. The takeover time is not 
defined for $\beta=1$. The selective pressure drops 
when the value of $\beta$ increases. We can see on figure \ref{growthPC} the 
growth of the number of copies of the best solution in the population (top) and its growth rate (bottom). There are two stages in the shape of the curve. The growth rate is linear in the first part and quadratic in the second part.
When using this selection scheme, the diffusion of the best solution is still isotropic. 
So the best solution roughly propagates describing an obtuse square as long as no side 
of the grid is reached. This corresponds to the first part of the growth rate curve. 
Once the sides are reached by some copies of the best solution, the dynamic changes as we 
can observe on the second part of the growth rate curves.

\subsection{Punctuated equilibria model}

\begin{figure}[ht!]
\begin{center}
\rotatebox{270}{\includegraphics[width=6cm,height=6cm]{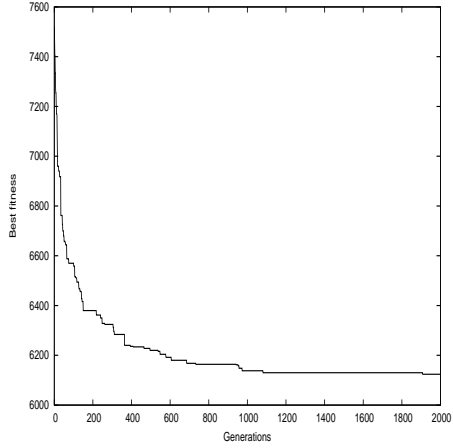}} 
\end{center}
\caption{Example of evolutionary run}
\label{run}
\end{figure}

In this section, we propose a new model which will help in the understanding of the search dynamics 
of an Evolutionary Algorithm. This model was first designed for a cellular EA but can be easily extended 
to any kind of evolutionary algorithm. We consider a cEA initialized with random solutions. We make sure that the best solution in the population is unique.
 Our goal is to simulate an 
evolutionary run: We simulate recombination 
 and mutation operators with probabilities that the mating is efficient or not (i.e. produces a new best solution). We consider three different types of matings: between two copies of 
the best solution (mating $11$), between one copy of the best solution and 
one sub-optimal solution (mating $01$) and between two sub-optimal solutions 
(mating $00$).
 We introduce probabilities $P_{11}$, $P_{01}$ and $P_{00}$ that matings of type $11$, $01$ and $00$ produce a new best solution, fitter than the previous 
best one. Figure \ref{run} is an example of evolutionary run on some optimization 
problem (minimisation task). We can see that there are some stagnation periods 
where the best solution don't improve. Then, an amelioration occurs and the 
population enters another stability period. An evolutionary run is a 
sum of stagnation periods and punctual improvements. Our punctuated equilibria 
model computes the probability of improving the best solution in the population 
according to the variables described above.

With this model, 
the probability of finding a new best solution at a given generation $t$ is :  
$$ p(t) = 1 - (1-P_{00})^{n_{00}(t)}(1-P_{01})^{n_{01}(t)}(1-P_{11})^{n_{11}(t)}$$
where $n_{00}(t)$, $n_{01}(t)$ and $n_{11}(t)$ are the number of matings of each type for the generation $t$. 

The average time to find a new best solution is given by :
$$E = \sum_{t \geq 1} t p(t)$$
%Assuming that $P_{00}$, $P_{01}$ and $P_{11}$ are constant, 
The performance of an algorithm can be measured by the time $E$ needed to 
find a new best solution but also by the probability $P$ of improvement in 
a preset time $T$. 
We have the probability of improving the best solution in $T$ generations : 

$$
\begin{array}{rcl}
P & = & 1 - \prod_{t=1}^T ( 1 - p(t) ) \\

P & = & 1 - ( 1 - P_{00} )^{\Sigma_{00}(T)}  ( 1 - P_{01} )^{\Sigma_{01}(T)}  ( 1 - P_{11} )^{\Sigma_{11}(T)} \\
\end{array}
$$
with $\Sigma_{ij}(T) = \sum_{t=1}^{T} n_{ij}(t)$ the sum over $T$ of mating of each type.

The parameters $P_{ij}$ are problem dependent
and the values of $\Sigma_{ij}$ are given by the selection scheme used. 
The selection process is usually controlled by a parameter such as the tournament size or in the case of the CS: $\beta$.
This parameter should be used to maximize the probability\footnote{In the following equations, we only denote the 
dependance on $\beta$ for $P$ and $\Sigma_{ij}$ for readability.} $P$.
Intuitively, the ideal selection process maximizes the $\Sigma_{ij}$ which have the higher $P_{ij}$.
More precisely, 
assuming that the control parameter of the selection process is $\beta$, 
the parameter $\beta^{*}$ which maximizes the probability $P(T)$ verifies:

$$
\begin{array}{rcl}
\frac{dP}{d\beta}(\beta^{*}) & = & 0 \\
\end{array}
$$

which gives:

\begin{equation}
\label{eq-optimal}
\begin{array}{ll}
& \log ( 1 - P_{00} ) \frac{\partial \Sigma_{00}}{\partial\beta}(\beta^{*}) \\
+ & \log ( 1 - P_{01} ) \frac{\partial \Sigma_{01}}{\partial\beta}(\beta^{*}) \\ 
+ & \log ( 1 - P_{11} ) \frac{\partial \Sigma_{11}}{\partial\beta}(\beta^{*}) = 0 \\
\end{array}
\end{equation}

If it is possible to have a model of $\Sigma_{ij}(\beta)$,
it would be possible to calculate the optimal $\beta$ as a function of $P_{ij}$.

In this model,
the exploration/exploitation tradeoff is given by the number of each
possible matting ($00$, $01$ and $11$).
The model could be used to explain the probability and the time to
find a new best solution according to the selective pressure,
and also to tune the value of parameters which have an impact on the 
selective pressure,
such as $\beta$,
to have the highest probability to evolve toward a new best solution.
Equation \ref{eq-optimal} gives precisely the best
exploration and exploitation tradeoff and allows computing the optimal
value of $\beta$ (in our case) for this trade-off. In the following, we will show
the validity of the PEM on some optimization problems.

\section{QAP and NK landscapes}

In this section, we study the effect of selective pressure on 
performances through experiments of a cEA with CS 
on two well-known classes of problems. The optimal exploration / exploitation 
tradeoff found will be explained thanks to the PEM 
presented in the previous section. 

\subsection{Problems presentation}

The problems proposed, Quadratic Assignment Problem and NK landscapes,
are known to be difficult to optimize.
The important number of instances of the Quadratic Assignment Problem and 
the tunable parameters of the NK landscapes allow managing the difficulty 
of the problems.

\subsubsection{Quadratic Assignment Problem}

This section presents the Quadratic Assignment Problem (QAP) which is known to be difficult to optimize.
The QAP is an important problem in theory and practice as well. It was introduced by Koopmans and Beckmann
in 1957 and is a model for many practical problems \cite{Koopmans57}. 
The QAP can be described as the problem of assigning a set of facilities to
a set of locations with given distances between the locations and given flows between the 
facilities. The goal is to place the facilities on locations in such a way that the sum
 of the products between flows and distances is minimal.
\linebreak
Given $n$ facilities and $n$ locations, two $n \times n$ matrices $D=[d_{kl}]$ and $F=[f_{ij}]$
 where $d_{kl}$ is the distance between locations $k$ and $l$ and $f_{ij}$ the flow between 
 facilities $i$ and $j$, the objective function is : \\
\begin{displaymath}
\Phi = \sum_{i}\sum_{j}d_{p(i)p(j)}f_{ij}
\end{displaymath}
where $p(i)$ gives the location of facility $i$ in the current permutation $p$.
\linebreak
 Nugent, Vollman and Ruml proposed a set of problem instances of different sizes noted for their difficulty \cite{Nugent68}. The instances they proposed are known to have multiple local optima, so they are difficult for an evolutionary algorithm.
The best algorithm known is the fast hybrid evolutionary algorithm
\cite{Mise06} which combines an evolutionary algorithm with an
improvement of the fast tabu search of Taillard.

\subsubsection*{Set up}

We use a population of 400 solutions placed on a square grid ($20\times 20$). Each solution
 is reprensented by
 a permutation of $N$ where $N$ is the size of a solution.
 The algorithm uses a crossover that preserves the permutations:
\begin{itemize}
\item
  Select two solutions $p_1$ and $p_2$ as genitors.
\item
  Choose a random position $i$.
\item
  Find $j$ and $k$ so that $p_1(i) = p_2(j)$ and $p_2(i) = p_1(k)$.
\item
  exchange positions $i$ and $j$ from $p_1$ and positions $i$ and $k$ from $p_2$.
\item
  repeat $N/3$ times this procedure where $N$ is the size of an solution.
\end{itemize}

This crossover is an extended version of the UPMX crossover proposed in \cite{Migkikh}.
The mutation operator consists in randomly selecting two positions from the solution
 and exchanging those positions. The crossover rate is 1 and we do a mutation per solution.
We perform 200 runs for each tuning of the two selection operators. An elitism replacement 
procedure guarantees the solutions stay on the grid if they are fitter than their 
offspring.

\subsubsection{NK landscapes}

The NK landscapes were proposed by Kaufmann to
model the boolean network and used in optimisation in order to explore how epistasis is linked to the ruggedness of 
search spaces \cite{Kauffman93}. Epistasis corresponds 
to the degree of interactions between the ``loci'' of a solution and ruggedness is the number of local optima of the search space.
 The main characteristic of NK Landscapes is that they allow tuning the epistasis level with a single parameter $K$. 
The parameter $N$ determines the length of the solutions.

The fitness of solutions for a NK landscape is given by the function $$ f : \lbrace 0,1 \rbrace ^N \rightarrow [0,1] $$ defined on binary strings of length $N$.
Each binary string is a solution with $N$ locations. 
An \textit{atom} with fixed epistasis level is represented by a fitness component $$ f_i : \lbrace 0,1 \rbrace ^{K+1} \rightarrow [0,1] $$ associated to each
 bit $i$. It depends on the value of the bit $i$ and on the value of $K$ other bits of the string ($K$ must fall between $0$ and $N-1$).
The fitness $f(x)$ of $x \in \lbrace 0,1 \rbrace ^N$ is the average of the values of the $N$ fitness components $f_i$ :
$$ f(x) = \frac{1}{N}\sum_{i=1}^N f_i(x_i,x_{i1},...,x_{ik})$$
where $ \lbrace i_1,...,i_k \rbrace \subset \lbrace 1,...,i-1,i+1,...,N \rbrace $. Many ways have been proposed to choose the $K$ other locations from the $N$ of the
 solutions. The mainly used ones are adjacent and random neighborhoods. With the first one, the $K$ nearest locations of the location $i$ are chosen (the solution is taken
 to have periodic boundaries). With the random neighborhood, $K$ locations are randomly selected from the solution. Each fitness component $f_i$ is specified by extension,
 \textit{ie} a random number $y_{i,(x_i,x_{i1},...,x_{ik})}$ from $[0,1]$ is associated with each element $(x_i,x_{i1},...,x_{ik})$ from $\lbrace 0,1 \rbrace^{K+1}$.
Those numbers are uniformly distributed in the interval $[0,1]$.\\ 

\subsubsection*{Set up}

The size of the population is $400$ solutions. The crossover used is a one point crossover, applied with 
a probability of $1$. The mutation is a bit flip applied with a probability of $\frac{1}{n}$ where $n$ is 
the size of a solution. We perform $200$ runs for every parameter set, and each run stops after $1500$ generations. 
Runs are performed on instances of sizes $N=32$, with $K \in { 2..12 }$. 

\subsection{Performances}

\begin{figure}[ht!]
\begin{center}
\rotatebox{270}{\includegraphics[width=6cm,height=6cm]{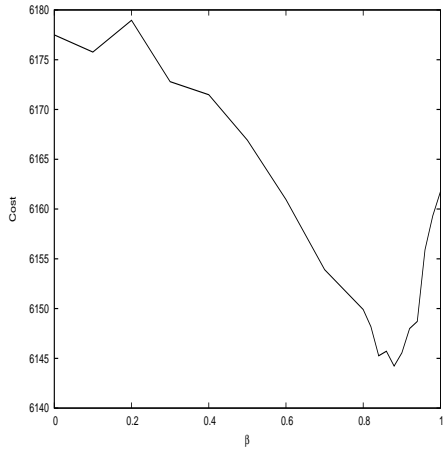}} 
\end{center}
\caption{Performances on nug30 for a cEA using CS.}
\label{nug30perf}
\end{figure}

Figure \ref{nug30perf} and table \ref{otherQAP} show performances of a 
cEA using CS on some QAP  instances 
of various sizes. The instance in figure \ref{nug30perf} is a well-known 
instance of size $30$. 
The first fact that we notice when looking at these results is that there is an optimal 
setting, different from the extreme values $0$ and $1$ for the parameter $\beta$. This indicates 
that for a certain setting of the parameters, and thus for a certain selective pressure, 
the search dynamic leads to optimal results. Curves representing the instances 
summarized in table \ref{otherQAP} have the same shape as figure \ref{nug30perf}.
 On each instance, the optimal value of $\beta$ is around $0.85$.
The performances increase up to these values and then decrease. 
Performances of CS are significantly better than the one obtained with a cEA using standard binary tournament selection. The standard cEA is observable 
on the curve at the points $\beta = 0.2$ and is reported in the table \ref{otherQAP}. We can also notice in table \ref{otherQAP} that the standard deviation 
is lower for the optimal value of $\beta$ than with a cEA with binary tournament
 selection.

\begin{table}
\begin{center}
\caption{Avg. results and std.dev.\newline on QAP instances}
\label{otherQAP}
\begin{tabular}{|c|c|c|c|}
\hline
Instance & Std cGA & Best avg. results & Opt. $\beta$ \\
\hline
Nug30 & $6178_{[ 28 ]}$ & $6144_{[ 14 ]}$ & $0.88$ \\
Tai40a & $3.23 \times 10^{6}_{[ 14343 ]}$ & $3.21 \times 10^{6}_{[ 12000 ]}$ & $0.84$ \\
Sko42 & $15969_{[ 75 ]}$ & $15909_{[ 34 ]}$ & $0.82$ \\
Tai50a & $5.092 \times 10^{6}_{[ 20721 ]}$ & $5.080 \times 10^{6}_{[ 13372 ]}$ & $0.82$ \\
Tai60a & $7429118_{[ 27760 ]}$ & $7385390_{[ 19391 ]}$ & $0.86$ \\
\hline
\end{tabular}
\end{center}
\end{table}

\begin{figure}[ht!]
\begin{center}
\rotatebox{270}{\includegraphics[width=6cm,height=6cm]{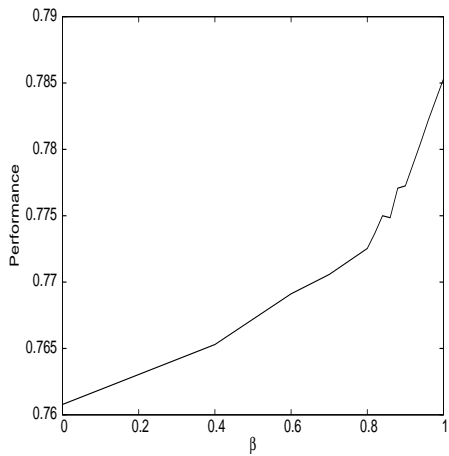}} 
\end{center}
\caption{Performances on NK with $N=32$ and $K=10$ for a cEA using CS.}
\label{nk32-10}
\end{figure}

Figure \ref{nk32-10} and table \ref{otherNK} present performances of a cEA with CS on some instances of NK landscapes. Parameters of the landscapes are $N=32$
 and $K=10$ for the figure \ref{nk32-10} and are summarized in table 
\ref{otherNK} for the other instances.

 We can see that the shape of the 
performances' curve is different from the QAP curve. The performance increases until $\beta$ reaches its maximum value. The same results are obtained for 
all the instances in table \ref{otherNK}. The parameter $K$ tunes the difficulty
 of the instance. We can see that for $K=2$, there is no optimal value for 
$\beta$. The reason is that the optimum is always found. For $K=4$, the standard cEA sometimes get stuck in a local optimum, and with $\beta=1$ our algorithm 
always find the optimum. On every instance, except $K=2$, the optimal value for $\beta$ is $1$.   

However, this value $\beta=1$ is a particular one, since it breaks all 
communications on the grid. 
As long as the value of the parameter increases, the chances 
of selecting two different solutions for recombination decrease.
 For $\beta = 1$, the algorithm is the parallelisation of as much hill 
climbers as there are cells on the grid : It constantly selects the center cells 
of the neighborhoods, so there is no crossover and any amelioration 
is due to a bit flip. 

So the best setting for CS can be compared to the parallelisation of as many 
hill climbers as there are cells on the grid. The parallelisation of hill climbing seems to be a good algorithm for 
solving NK landscapes problems, which could be explained by the size of basins 
of attraction \cite{Verel08} and \cite{Verel08b}.

\begin{table}
\begin{center}
\caption{Avg. performances and std.dev.\newline on NK instances with $N=32$}
\label{otherNK}
\begin{tabular}{|c|c|c|c|}
\hline
$K$ & Std cGA & Best avg. results & Best $\beta$ \\
\hline
$2$ & $0.734329_{[ 0 ]}$ & $0.734329_{[ 0 ]}$ & $[ 0,1 ]$ \\
$4$ & $0.79597_{[ 0.003 ]}$ & $0.798197_{[ 0 ]}$ & $1$ \\
$6$ & $0.782934_{[ 0.01 ]}$ & $0.799124_{[ 0.003 ]}$ & $1$ \\
$8$ & $0.771277_{[ 0.01 ]}$ & $0.789103_{[ 0.004 ]}$ & $1$ \\
$10$ &$0.763510_{[ 0.01 ]}$ & $0.785115_{[ 0.003 ]}$ & $1$ \\
$12$ & $0.750043_{[ 0.01 ]}$ & $0.774479_{[ 0.009 ]}$ & $1$ \\
\hline
\end{tabular}
\end{center}
\end{table}

\subsection{Probabilities of discovering better solutions}

In order to explain the optimal values of $\beta$ for QAP and NK-Landscape
 and to validate the PEM, 
we compute $P$, the probability of discovering a new best solution 
in the population taken from the PEM, with 
real data. We calculated it for one instance of QAP and one instance of 
NK-Landscapes. With this calculation we want to find the value of $\beta$
 that maximizes the probability of discovering a new solution. 
This probability depends 
on the value of $\Sigma_{ij}$, and thus on time: if at a generation 
$t$ no new solution is discovered, the actual best solution spreads in the 
population according to the selective pressure. If during an interval of time 
corresponding to the takeover time no new solution is discovered then 
 the population converges. We can compute the ideal $\beta$ value for a given 
number of generations $T$ because $\Sigma_{ij}( T )$ relies on $\beta$ and on time: after $T$ generations $\Sigma_{ij}( T )$ is different according to 
$\beta$, and for the optimal value of $\beta$ $\Sigma_{ij}( T )$ leads to the 
best probability $P$.

\begin{figure}[ht!]
\begin{center}
\rotatebox{270}{\includegraphics[width=6cm,height=6cm]{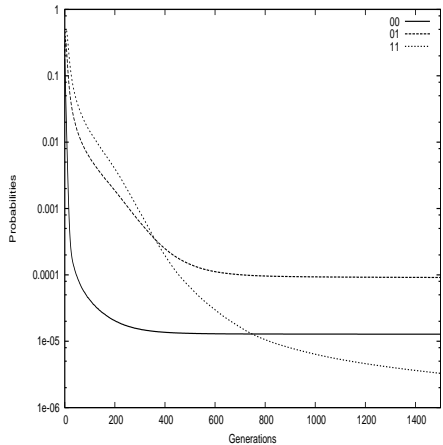}}
\end{center}
\caption{Estimated $P_{ij}$ on QAP instance nug30}
\label{nug30probas}
\end{figure}

We estimated the $\Sigma_{ij}$ with the same experiments done to 
compute growth curves and takeover time. We averaged the number of matings of 
each type at each generation over $10^{3}$ runs.  Then, we needed to know the 
probabilities $P_{00}$, $P_{01}$ and $P_{11}$. We estimated these probabilities using 
a Bayesian process during the runs. We averaged the values obtained by 
generations over $500$ runs. Figure \ref{nug30probas} shows the result 
of the estimation of probabilities on the QAP instance Nug30. The ordonate scale
 is logarithmic because of the variations of probabilities. The curves 
representing the $P_{ij}$ intersect, so the value of $\beta$ which 
maximizes $P$ may change during a run. We computed $P$ with estimated values 
of $P_{ij}$ taken by steps of $50$ generations. The values of $\Sigma_{ij}$ 
are also generation dependent. For each value of $\beta$, we took 
the $\Sigma_{ij}$ value after $100$ generations: that is $\Sigma_{ij}(100)$.
 During a run, it would 
correspond to allowing a stagnation period of $100$ generations before 
stopping the run, which is reasonnable.

\begin{figure}[ht!]
\begin{center}
\rotatebox{270}{\includegraphics[width=6cm,height=6cm]{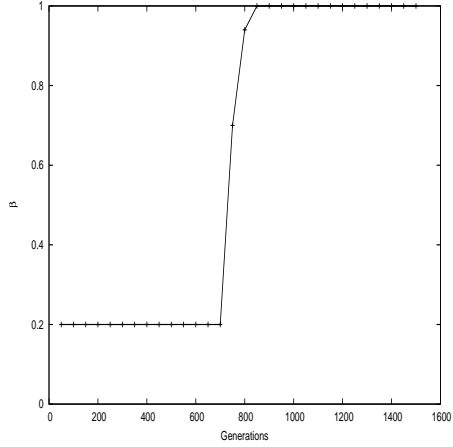}}
\end{center}
\caption{Optimal values of $\beta$ on QAP instance nug30}
\label{nug30opt}
\end{figure}

\begin{figure}[ht!]
\begin{center}
\rotatebox{270}{\includegraphics[width=6cm,height=6cm]{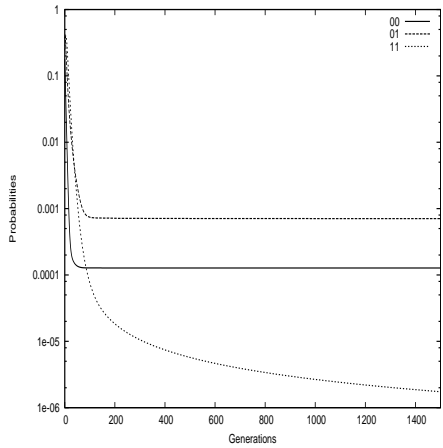}}
\end{center}
\caption{Estimated $P_{ij}$ on NK-Landscape with $N=32$ and $K=10$}
\label{nkprobas}
\end{figure}

\begin{figure}[ht!]
\begin{center}
\rotatebox{270}{\includegraphics[width=6cm,height=6cm]{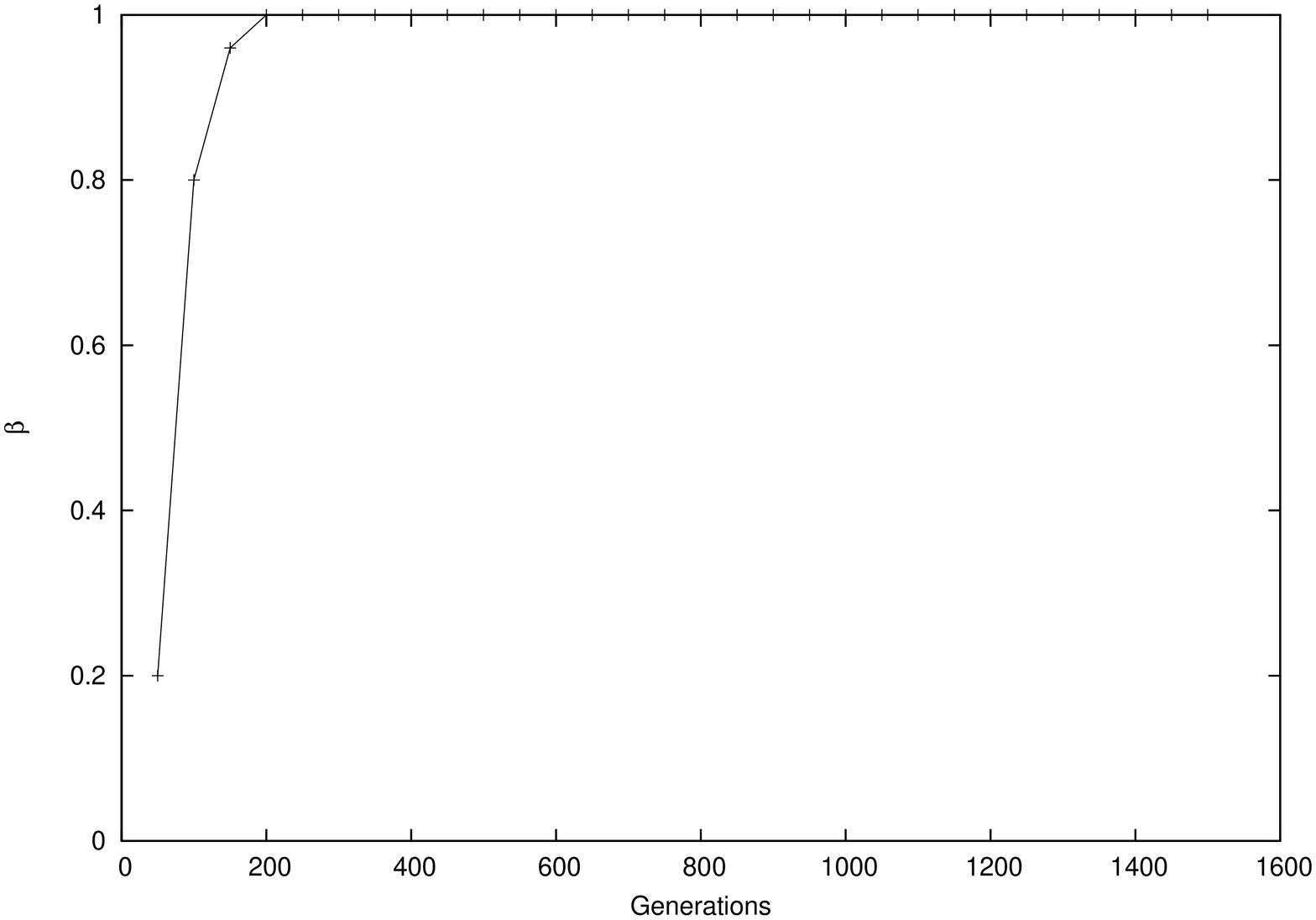}}
\end{center}
\caption{Optimal values of $\beta$ on NK-Landscape $N=32$, $K=10$}
\label{nkopt}
\end{figure}

The figure \ref{nug30opt} shows the optimal values of $\beta$ as a function of 
generations for the QAP instance Nug30. During the first $700$ generations, 
the optimal value is $\beta=0.2$. Then, there is a transition of approximatively
$150$ generations. During this phase, the ideal value of $\beta$ grows 
until it reaches $1$. In our experiments, we observe optimal values of $\beta$
 between $0.8$ and $0.9$ according to the QAP instance. Values of $\beta$ are 
constant during the runs. But the PEM shows that the selective pressure should 
be strong at the beginning (low values of $\beta$) and then weak (high values 
of $\beta$).  If $\beta$ is constant, intermediate values  
in the range $[ 0.8, 0.9 ]$ give the best average selective 
pressure for QAP instances.

The figure \ref{nkopt} shows the optimal values of $\beta$ as a function of 
generations for a NK-Landscape with $N=32$ and $K=10$. We can see that the 
optimal $\beta$ value increases fastly and reaches $1$ in the early generations.
The ideal selective pressure is weak, and it is not surprising that the best 
performances are obtained when $\beta=1$ in our experiments. Figure \ref{nkprobas} shows the estimation of $P_{ij}$ as a function of generations. We can see 
that the curve representing $P_{11}$ drops down very fast. With a negligible 
probability of improving the current best solution with mutations, there 
is no sense in spreading this solution. With $\beta=1$, the current best 
solution in the population cannot spread. 

The PEM has been used in order to explain the exploration / exploitation 
trade-off on two different classes of problems. Coupled with the centric 
selection, it showed the ideal selective pressure along the 
search process. This model can be used to tune any parameter which has 
some influence on the number of matings of each type defined in the previous
 section. The computation cost is low, since the estimation of probabilities 
by a Bayesian process is precise: we averaged the estimation on $500$ runs 
but the standard deviation was low ($\approx 10^{-6}$).The $\Sigma_{ij}$ are 
only computed once since they are independent from the optimization problem 
tackled. 

\clearpage

\section{Conclusion}

The exploration/exploitation trade-off is an 
important issue in evolutionary algorithms.
 In this paper, we propose a model that 
takes into account stochastic variations  and improvement 
of the quality of the solutions, the punctuated equilibria model.
 In order to study the exploration / 
exploitation trade-off we propose a tunable selection operator: the 
centric selection. By monitoring the probability of selecting the center cell 
of neighborhoods for a tournament selection, this selection operator allows 
tuning accurately and continuously the selective pressure with one single 
parameter ($\beta$).  
The performance results on QAP instances 
and NK-Landscapes showed different optimal settings of the centric selection,
 and thus different ideal selective pressures. Using the punctuated 
equilibria model, we put in evidence the optimal values of the centric 
selection's control parameter observed on QAP instances and 
NK-Landscapes. The punctuated equilibria model also put in evidence that 
the ideal selective pressure is not constant during the search process 
in the case of QAP instances.   

In this paper, we used the PEM in order to explain experimental results. 
In future works, we will use it in order to predict optimal exploration / 
exploitation trade-offs and to adapt the selective pressure during the runs.
To do so, we will both estimate the $P_{ij}$ and tune the selection 
operator online during the search process. The centric selection will 
be used in auto-adaptative algorithms with the advantage of modifying 
the exploration / exploitation ratio with a single parameter. It will also 
be applied on real problems and compared to other optimization methods.

\end{document}